\documentclass{article}
\usepackage{spconf,amsmath,amsfonts,amsthm,booktabs,multirow,algorithm,algpseudocode}
\usepackage{graphicx}
\usepackage{color}

\usepackage{textalpha}

\newcommand{\fop}{\mathcal F}

\newcommand{\R}{\mathbb R}
\newcommand{\Rpos}{\R_{\geq0}}
\newcommand{\dataset}{\mathcal Y}
\newcommand{\datapoint}{y}

\newcommand{\struct}{x}
\newcommand{\template}{\struct_0}

\newcommand{\reg}{\mathcal R}
\newcommand{\regparam}{\lambda}

\newcommand{\loss}{\mathcal L}
\newcommand{\reconspace}{\R^{3\times\natoms}}
\newcommand{\ndata}{M}
\newcommand{\natoms}{N}

\newcommand{\graph}{G}
\newcommand{\nn}{f}

\newcommand{\planepoint}{q}
\newcommand{\psf}{h}
\newcommand{\ctf}{\hat \psf}

\title{Protein Graph Neural Networks for Heterogeneous Cryo-EM Reconstruction}
\name{Jonathan Krook$^{\dagger \star}$, Axel Janson$^{\dagger \star}$, Joakim Andén$^{\dagger}$, Melanie Weber$^{\, \ddagger}$, Ozan Öktem$^{\dagger}$ \thanks{MW acknowledges support from NSF award CBET-2112085, an Alfred P. Sloan Fellowship, an Aramont Fellowship for Emerging Science Research, as well as Schmidt Sciences (Grant G-25-69786). JK and OÖ acknowledge support from Swedish Research Council grant 2020-03107. AJ and JA acknowledge support from Swedish Research Council grant 2023-04143. The computations were run on the FASRC Cannon cluster supported by the FAS Division of Science Research Computing Group at Harvard University.
}}
\address{\qquad $^{\dagger}$KTH Royal Institute of Technology \qquad\qquad\qquad\qquad\qquad $^{\ddagger}$Harvard University \qquad\qquad\qquad\qquad \\\qquad Department of Mathematics \qquad\qquad\qquad\quad School of Engineering and Applied Sciences\qquad}

\begin{document}

\maketitle

\def\thefootnote{$\star$}\footnotetext{These authors contributed equally to this work.}\def\thefootnote{\arabic{footnote}}

\begin{abstract}
We present a geometry-aware method for heterogeneous single-particle cryogenic electron microscopy~(cryo-EM) reconstruction that predicts atomic backbone conformations. To incorporate protein-structure priors, we represent the backbone as a graph and use a graph neural network~(GNN) autodecoder that maps per-image latent variables to 3D displacements of a template conformation. The objective combines a data-discrepancy term based on a differentiable cryo-EM forward model with geometric regularization, and it supports unknown orientations via ellipsoidal support lifting~(ESL) pose estimation. On synthetic datasets derived from molecular dynamics trajectories, the proposed GNN achieves higher accuracy compared to a multilayer perceptron~(MLP) of comparable size, highlighting the benefits of a geometry-informed inductive bias.
\end{abstract}
\begin{keywords}
graph neural networks, cryo-EM, atomic modeling, autodecoders, inverse problems
\end{keywords}

\begin{figure*}[t]
    \centering
    \begin{minipage}[t]{\linewidth}
        \includegraphics[width=\linewidth]{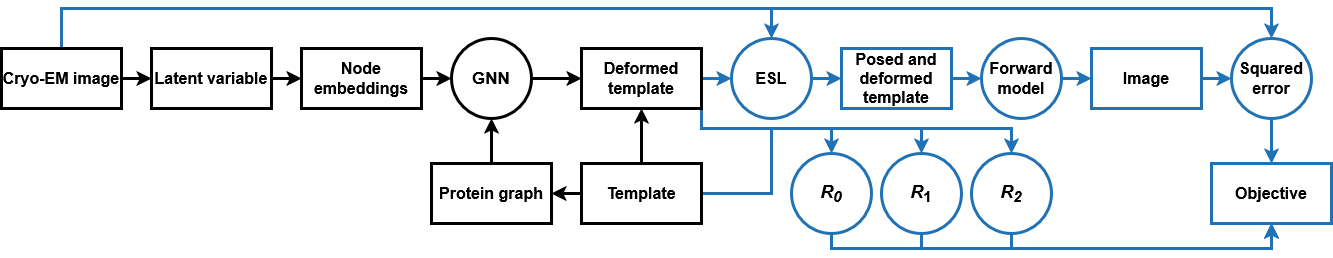}
    \end{minipage}
    \caption{Principal overview of our method. A data point in the form of an EM image is indexed by a low dimensional variable, and then mapped via a GNN to a reconstructed conformation. Blue indicates components that are only used during optimization, where the deformed template is posed and compared to data, and regularization is added.}
    \label{fig:lossfunction}
\end{figure*}

\section{Introduction}
Characterizing the structure and dynamics of biological macromolecules is of fundamental importance in a variety of research fields, such as cell and molecular biology, drug design, and understanding the underlying mechanism behind various diseases. 
This is particularly true for proteins -- long chains of amino acids which fold into complex shapes -- whose function is largely determined by the various structural configurations, known as \emph{conformations}, it can take on.
Proteins can change between these conformations on their own, in response to a change in its environment, or in the presence of another molecule bound to a particular location.
As such, the protein works as a \emph{molecular machine}, going through various states which results in a certain outcome.

A popular technique for elucidating these conformations is single-particle cryogenic electron microscopy~(cryo-EM).
Here, a large number of copies of the same molecule are purified into a sample solution, which is then rapidly frozen to prevent the formation of ice crystals.
The sample is then imaged in a transmission electron microscope, resulting in large images known as micrographs.
Each micrograph contains hundreds of particle projections, each representing the molecule in a different conformation and viewed from a different orientation.
By extracting single-particle images, we obtain datasets of hundreds of thousands of images representing a range of conformations.

Three important challenges arise when trying to extract the 3D atomic configurations from these particle images.
The first is that the images have a very high noise level.
Since large electron doses will damage the sample, the dose is kept low, but this in turn reduces the signal-to-noise ratio~(SNR). This problem is exacerbated in the case of small molecules, where the signal energy is already quite low.
The second challenge is that the poses (3D orientations and 2D in-plane offsets) of the particles are not known in advance, but must be estimated.
Finally, most reconstruction methods produce an estimate of the 3D volumetric potential of the molecule, into which an atomic model must then be fitted in a process known as \emph{model building}~\cite{Alnabati2020}.
Consequently, any errors introduced due to the low SNR and unknown orientations are amplified when trying to align the atomic models to the reconstructed potential.

In this work, we consider the 3D atomic reconstruction problem for datasets exhibiting \emph{continuous heterogeneity}, i.e., where the conformations are sampled from a continuous distribution of conformations.
While several methods have been proposed to address the conventional (homogeneous) 3D reconstruction problem, accurate 3D reconstruction under continuous heterogeneity remains an open problem.


One popular approach has been to estimate the distribution of unknown states using variational autoencoders (VAEs).
These will take an individual image, encode it into a latent variable, which is then decoded into a 3D structure (either volumetric or as an atomic model).
The latent variable here becomes a proxy for the molecular conformation of the particle in that image.
This is then coupled with a standard projection-matching or expectation--maximization framework in order to estimate the orientations from the predicted 3D structure.
The most prominent examples of this line of work are given by Rosenbaum et al., 2021~\cite{rosenbaum2021inferring}, CryoDRGN~\cite{Zhong2021}, CryoDRGN-AI~\cite{Levy2025}, DynaMight~\cite{Schwab2024}, CryoSTAR~\cite{Li2024}, OPUS-DSD~\cite{Luo2023}, e2GMM~\cite{Chen2021}, and cryo-FIRE~\cite{Levy2022}.
While these methods have enjoyed significant success, they still suffer from a certain inaccuracy in their dependence on choice of architecture and optimization hyperparameters~\cite{Minkyu2024CryoBench}.

All of the 3D reconstruction methods listed above use MLP- and CNN-based architectures for their neural networks.
We posit that more specialized architectures, adapted to the protein geometry, have the potential to perform better.
To design such an architecture, we use a coarse-grained atomic model of a protein backbone chain to represent a 3D structure, similar to Rosenbaum et al., 2021~\cite{rosenbaum2021inferring}, and Chen \& Ludtke, 2021~\cite{Chen2021}.
As noted in these works, atomic models naturally incorporate geometric priors in the reconstruction method.
Further, we encode the prior information explicitly in the model by defining a graph neural network~(GNN) on the protein backbone chain.
While GNNs have been used previously to analyze protein data~\cite{Jamali2024}, they have not been used for 3D reconstruction in cryo-EM.

The contribution of this work is a new method for 3D reconstruction of cryo-EM datasets exhibiting continuous heterogeneity, using GNNs to represent the atomic backbone directly (see Fig.~\ref{fig:lossfunction} for an overview).
We validate our method on two synthetic datasets exhibiting continuous conformational heterogeneity, where significant gains are obtained compared to conventional MLP architectures.
Synthetic datasets provide access to ground truth conformations for validation, allowing us to compare the performance of different architectures.
These experiments are performed both in the case of known orientations and unknown orientations.
In the latter, we integrate the optimization procedure of the GNN with the ellipsoidal support lifting~(ESL) method in order to estimate the poses~\cite{Diepeveen2023}.
Our results show that the GNN architecture outperforms the MLP in terms of reconstruction accuracy, demonstrating the advantages of a compatible neural network architecture for modeling protein chains. 

This paper is organized as follows.
Section~2 describes the problem of heterogeneous reconstruction from cryo-EM data. Section~3 describes our proposed method, including the GNN autodecoder model which is its core component.
Finally, Section~4 presents our experiments on synthetic cryo-EM data.
\section{Problem Formulation}
Our aim is to reconstruct a set of $\ndata$ protein conformations from their indirect observations in the form of a dataset of cryo-EM images $\dataset=(\datapoint_i)_{i=1}^\ndata$. These images feature centered particles in different conformations, and their orientations are either known or unknown.
The sample conformations are assumed to exhibit continuous heterogeneity, meaning that there is continuous distribution of conformational states in the sample.
We represent a conformation by an $\natoms$-tuple of 3D coordinates $x=(x^1,\dots,x^\natoms)\in\reconspace$, where $\natoms$ is the number of amino acid residues in the protein being imaged, and the coordinates represent the $\text{C}_\alpha$-coordinates (every amino acid residue has one $\text{C}_\alpha$-atom and every third atom in a protein backbone chain is a $\text{C}_\alpha$-atom).
Thus, $x$ is a coarse-grained representation of the protein backbone, and not the complete atomistic conformation.
The goal is now to find a conformation $x_i$ corresponding to each image $y_i$.







\section{Method}
The idea is to optimize a neural network autodecoder with a GNN component, which deforms a template conformation into a predicted conformation for each image in the dataset.
This approach requires good pose estimates in order to accurately estimate the data discrepancy, for which we use a separate pose estimation scheme.


\subsection{Forward Model} \label{sec:forward}
Comparing reconstructed conformations with their corresponding 2D cryo-EM images during optimization requires access to a differentiable forward model $\fop$, which models TEM image formation ~\cite{Vulovic2013, Fanelli2008, vulovic2014}.

The input to the forward model is an $\natoms$-tuple of 3D coordinates, $x=(x^1,\dots,x^\natoms)\in\reconspace$, representing a protein conformation.
Here we describe the forward model assuming that the conformation is already posed -- the full forward model will be of the form $\fop(\phi.x)$, where $\phi \in \mathrm{SO}(3)$ is the pose acting on the conformation.
At each set of 3D coordinates, $x^i\in\R^3$, an isotropic Gaussian is placed, approximating the potential corresponding to each amino acid residue.
The particles represent unstained proteins suspended in aqueous buffer, so they act as weak-phase objects in TEM imaging. 
A well-known approximation electron-specimen interaction is to model it by evaluating the ray transform of the aforementioned potential on lines parallel to the TEM optical axis (orthogonal to the imaging plane)~\cite{Oktem2015}. Since the ray transform maps Gaussians to Gaussians, this results in a 2D image $v_x\colon\R^2\to\R$ given as 
\begin{equation}
    v_x(\planepoint)=\sum_{i=1}^\natoms \frac{a_i}{2\pi\sigma_i^2}\exp\left(-\frac{\|\planepoint-Px^i\|_2^2}{2\sigma_i^2}\right),
\end{equation}
where $P\colon\R^3\to\R^2$ is the orthogonal projection operator mapping an arbitrary 3D point onto the 2D imaging plane that is orthogonal to the TEM optical axis.
Note the two sets of parameters $a_1,\dots,a_\natoms$, and $\sigma_1,\dots,\sigma_\natoms$, that depend on the properties of each amino acid residue~\cite{prins2024}.

The 2D image is finally convolved with a point spread function, $\psf$, that models the TEM optics, so $\fop(x)=\psf\ast v_x$. 
This point spread function is typically specified by its 2D Fourier transform $\ctf$, known as the contrast transfer function (CTF).
For the form of the CTF and its parameters, see the supplementary materials.

\subsection{Pose Estimation}
For pose estimation, we use a variant of the ESL method introduced by Diepeveen et al., 2023~\cite{Diepeveen2023}.
Recall that we assume that images are centered, so it remains to estimate the orientations.
We provide a summary of ESL and refer to the original paper for details.

Generally, pose estimation corresponds to a non-convex minimization problem, and finding a way to handle this non-convexity is a central challenge.
The approach adopted by Diepeveen et al., 2023~\cite{Diepeveen2023}, estimates poses for homogeneous reconstruction. 
First, it uses lifting to approximate and regularize the problem into one defined over discrete measures~$\mu^\star$ over $\operatorname{SO}(3)$. 
Further, this minimization problem is regularized to ensure that the optimal measure does not degenerate to a single point, yet is also localized in a small region.

Once the optimal measure has been calculated, the original paper proposes its barycenter as the optimal pose. 
In contrast, rather than taking a single pose as the best estimate, we compute the expected data discrepancy over the measure for reasons of efficiency.
In addition, to adapt ESL for heterogeneous reconstruction, our implementation runs the complete ESL algorithm separately for each predicted conformation. 
The result is one optimal measure~$\mu_i^\star$ per conformation. 
\subsection{GNN Autodecoder Model}
First, each data point $\datapoint_i\in\dataset$ is mapped to a corresponding latent variable $z_i\in\R^d$ using an index-based lookup table, similar to Edelberg and Lederman, 2023~\cite{Edelberg2023UsingVT}, which is then decoded by a neural network.
Accordingly, the network is an \emph{autodecoder}, not an autoencoder, since no formal encoder is included.
To understand why amortization is not required, note that the latent space is assumed to be low-dimensional, based on the hypothesis that the space of physically probable conformations has fewer degrees of freedom than the space of mathematically possible conformations. 
In addition, the number of images in a cryo-EM dataset is typically small enough to ensure that the total number of parameters in the lookup table is manageable. 

The neural network $\nn_\theta$ features a GNN acting on a protein graph $\graph$.
The graph $\graph$ is constructed from the template conformation $\template$ by introducing a node for each amino acid residue, and by placing an edge between any two nodes that are connected with either a peptide bond or a hydrogen bond from the secondary structure.
The point is that we build in prior information on which nodes should depend strongly on each other.

The first layer of $\nn_\theta$ is a linear layer mapping a latent variable $z$ to a set of node features $X^{(0)}$ on the graph $\graph$.
The node features are then transformed to a new set of node features $X^{(L)}$ via a $L$-layered GNN with skip connections.
Finally, the set of node features $X^{(L)}$ are mapped via a final linear layer acting node-wise, outputting a set of displacement vectors $\Delta\in\reconspace$, and the predicted conformations $\nn_\theta(z;G,\template)=\nn_\theta(z)=\template+\Delta$ are obtained by displacing the template conformation.
Model hyperparameters can be found in the supplemental materials.

Our GNN makes use of the graph convolutional layer by Kipf and Welling, 2017~\cite{kipf2017}.
Specifically, an input of node features $X^{\rm in}\in\R^{\natoms\times k}$ is transformed to output node features $X^{\rm out}\in\R^{\natoms\times s}$, defined at node~$i$ as
\begin{equation}
    X^{\rm out}_i=W^T\sum_{j\in\mathcal N(i)\cup\{i\}}(d_id_j)^{-1/2}X_j^{\rm in}+b,
\end{equation}
where $W\in\R^{k\times s}$ and $b\in\R^s$ are the layer weights and bias, respectively, while $\mathcal N(i)$ is the set of neighbors of node $i$, and $d_i=|\mathcal N(i)\cup\{i\}|$.

Thus, the graph convolutional layer aggregates information at each node from its neighbors.
Considering how we construct the protein graph $\graph$, we get that each layer aggregates information at each node from all bonded nodes.


\begin{algorithm}[t]
\caption{Optimization of autodecoder with ESL}
\label{alg:training}
\begin{algorithmic}[1]
\Require Images $\{y_i\}_{i=1}^M$, template $\template\in\mathbb{R}^{3\times N}$,
graph $G$,
regularization weights $(\lambda_0,\lambda_1,\lambda_2)$, 
batch size $B$
\State Initialize network weights $\theta$ and latents $\{z_i\}_{i=1}^M$.
\While{not converged}
    \State Sample index batch $\mathcal{B}\subset\{1,\dots,M\}$, $|\mathcal{B}|=B$.
    \State $\mathcal{L} \gets 0$.
    \For{$i \in \mathcal{B}$}
        \State $\hat{x}_i \gets  \nn_\theta(z_i; G, x_0)$ 
            \State $\mu_i^\star \gets \textsc{ESL}(y_i, \hat{x}_i)$ 
        \For{$\phi_i \in \mathrm{supp}(\mu_i^\star$)}
            \State $\mathcal{L} \gets \mathcal{L} + \mu_i^\star (\phi_i) \cdot \|y_i - \fop(\phi_i. \hat{x}_i)\|_2^2$
        \EndFor
        \State $\mathcal{L} \gets \mathcal{L} + \lambda_0 R_0(\hat{x}_i) + \lambda_1 R_1(\hat{x}_i) + \lambda_2 R_2(\hat{x}_i)$
    \EndFor
    \State Compute $\nabla_\theta \mathcal{L}$, $\{\nabla_{z_i}\mathcal{L}\}_{i\in\mathcal{B}}$ and update $\theta$, $\{z_i\}_{i\in\mathcal{B}}$.
\EndWhile
\State \Return Optimized decoder weights $\theta$ and latents $\{z_i\}$.
\end{algorithmic}
\end{algorithm}

\begin{figure}[t]
    \begin{minipage}[t]{.49\linewidth}
        \centering
        \includegraphics[trim=350 100 350 80, clip, width=4cm]{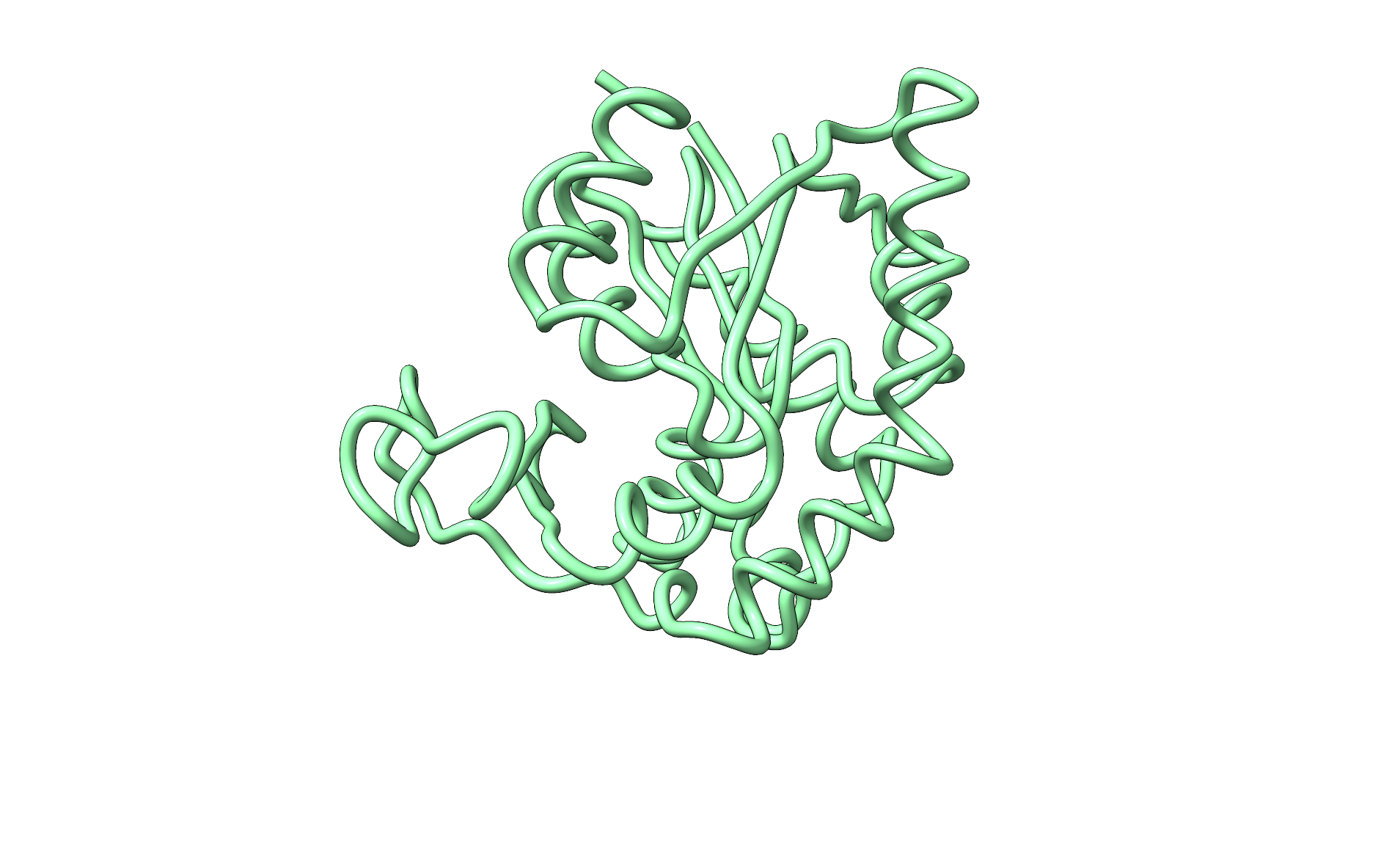}
        \includegraphics[trim=450 100 450 50, clip, width=4cm]{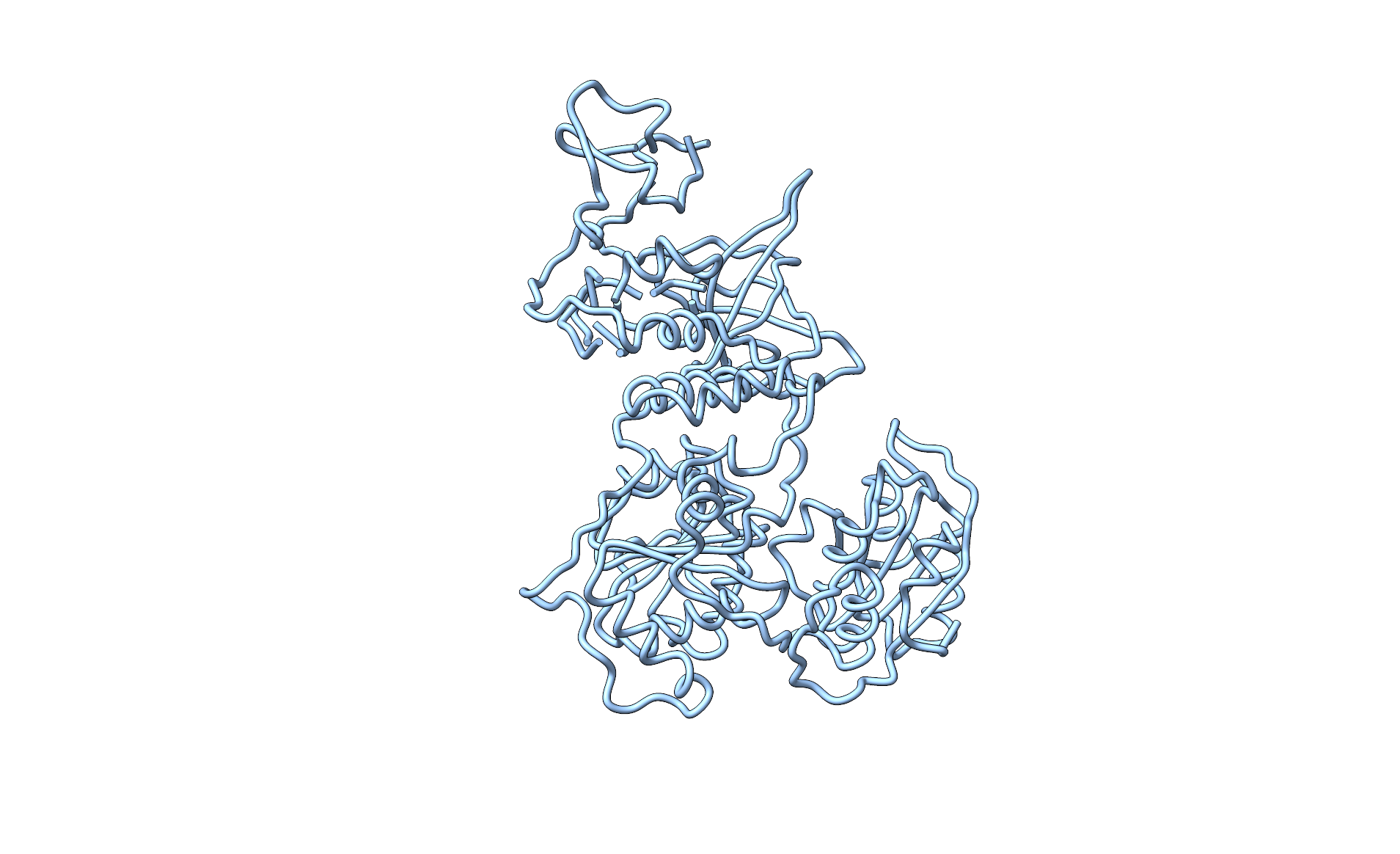}
    \end{minipage}
    \begin{minipage}[t]{.49\linewidth}
    \centering
        \includegraphics[trim=350 100 350 80, clip, width=4cm]{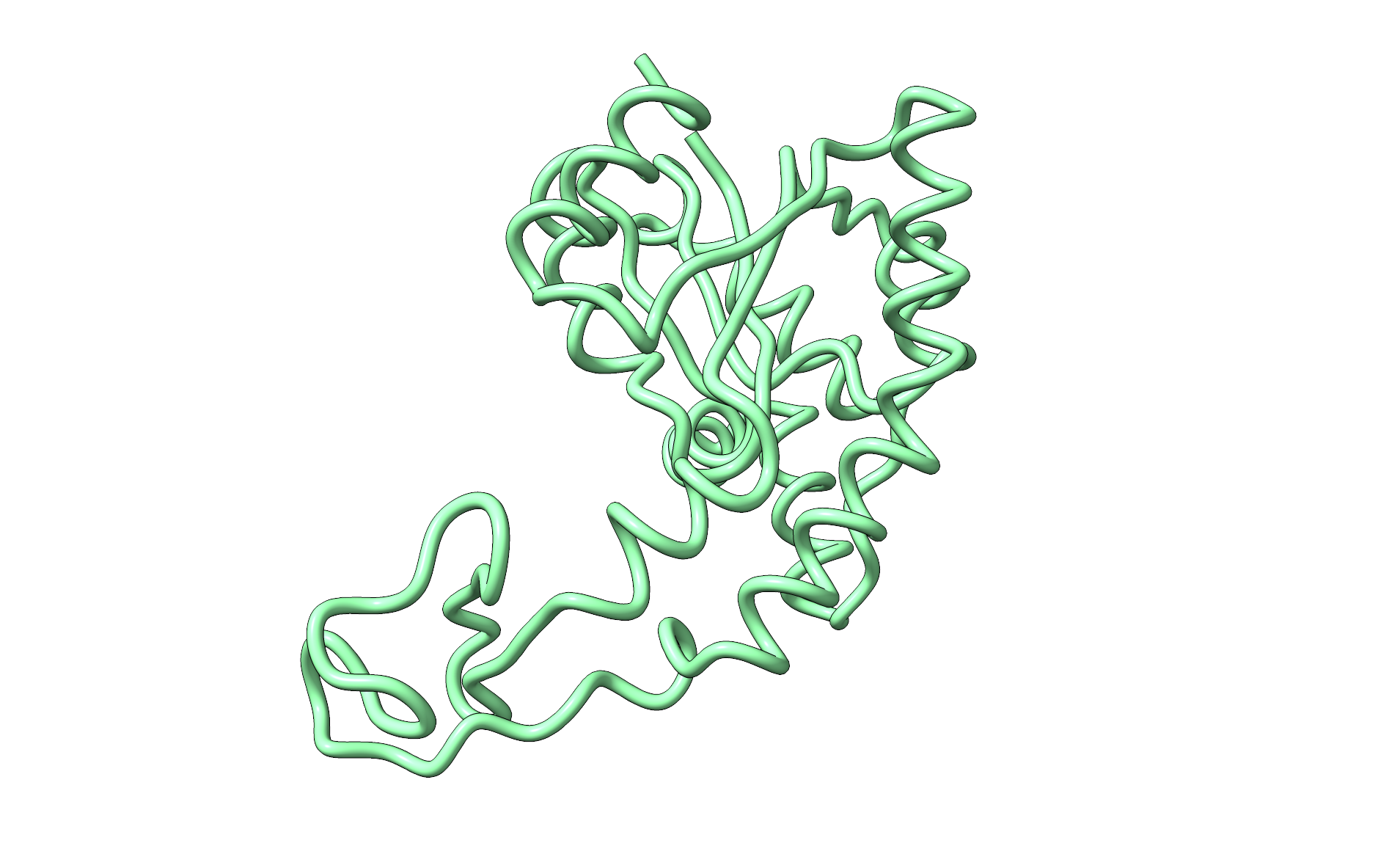}
        \includegraphics[trim=450 100 450 50, clip, width=4cm]{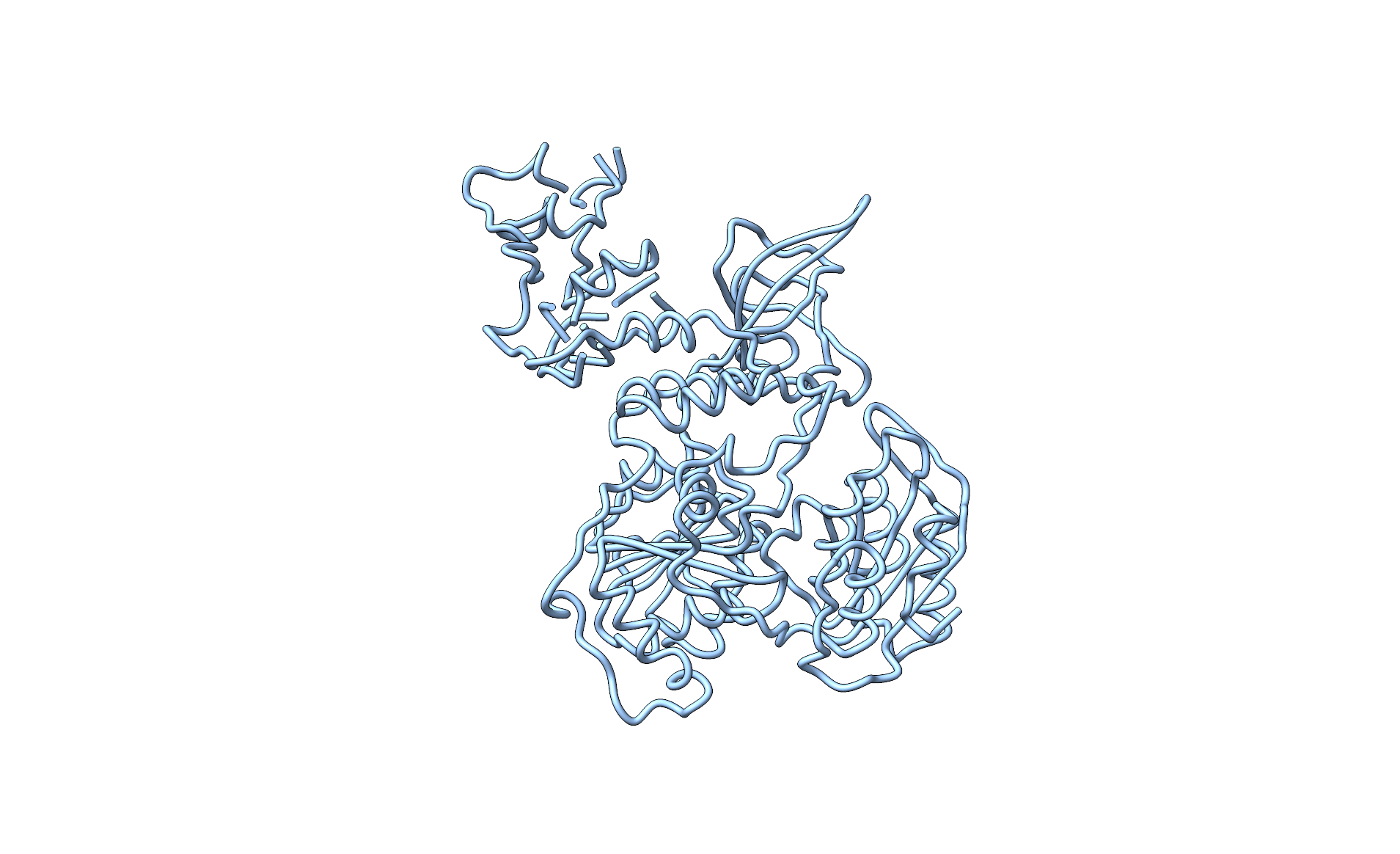}
    \end{minipage}
    \caption{Top: Closed and open conformations of ADK. Bottom: Heterogeneous conformations of NSP.}
    \label{fig:confs}
\end{figure}

\begin{figure}[t]
    \begin{minipage}[t]{\linewidth}
        \centering
        \centerline{\includegraphics[trim=10 60 0 40, clip, width=8.5cm]{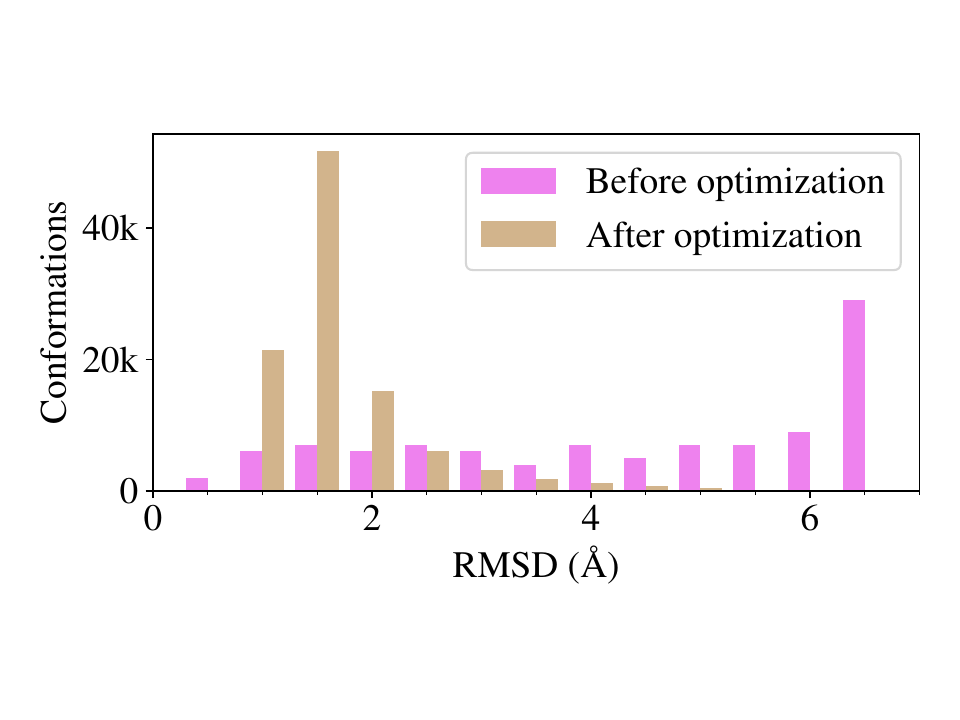}}
    \end{minipage}
    \caption{Histogram of ground truth RMSD values for all 102k conformations in the ADK dataset, before and after optimizing the GNN autodecoder with $\reg_2$-regularization and ESL.}
    \label{fig:histogram}
\end{figure}

\begin{figure*}[t] 
    \begin{minipage}[b]{0.24\linewidth}
        \centering
        \centerline{\includegraphics[trim=350 100 350 50, clip, width=0.64\linewidth]{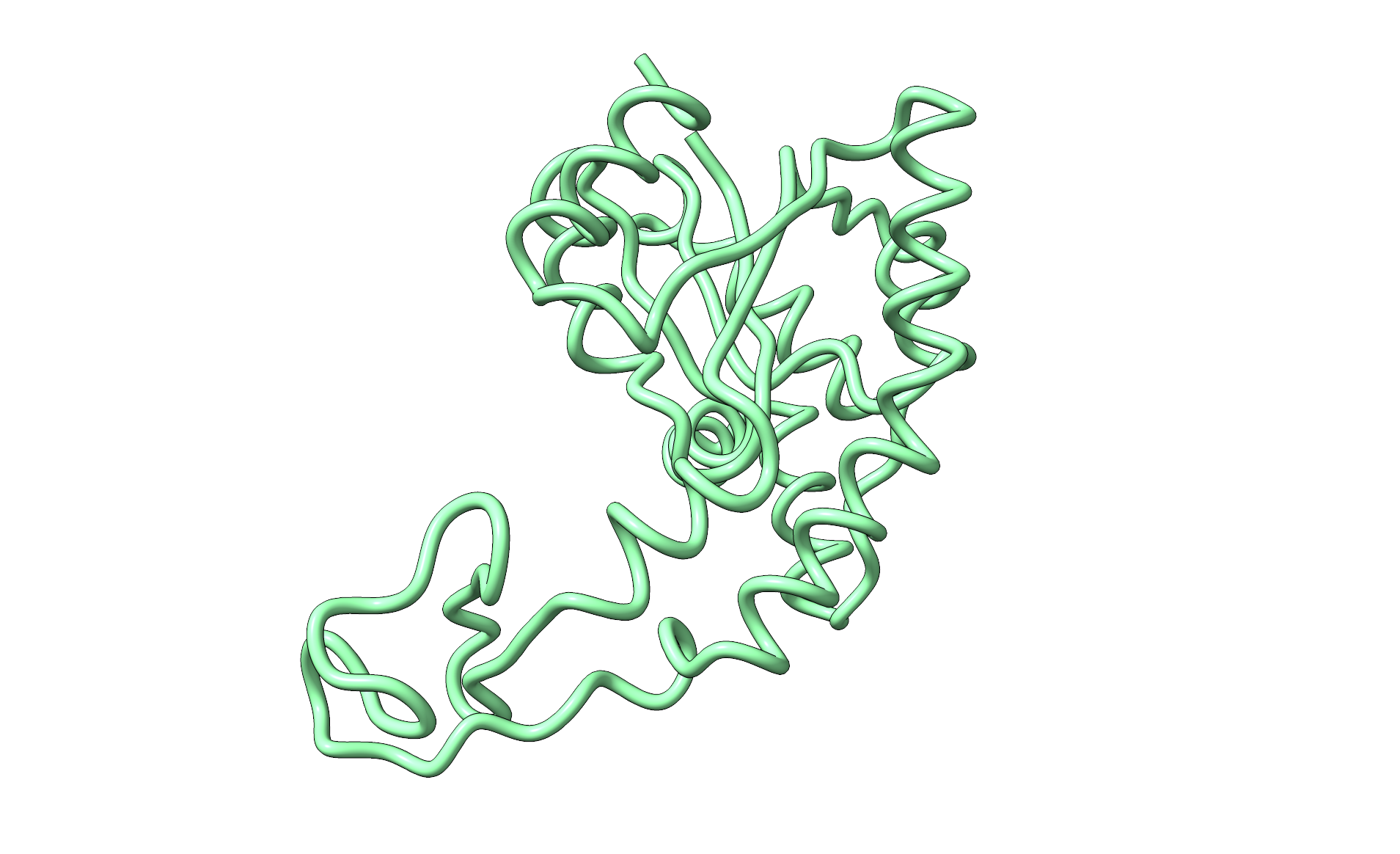}}
        \centerline{(a)}
    \end{minipage}
    \begin{minipage}[b]{0.24\linewidth}
        \centering
        \centerline{\includegraphics[trim=200 200 200 200, clip, width=0.64\linewidth]{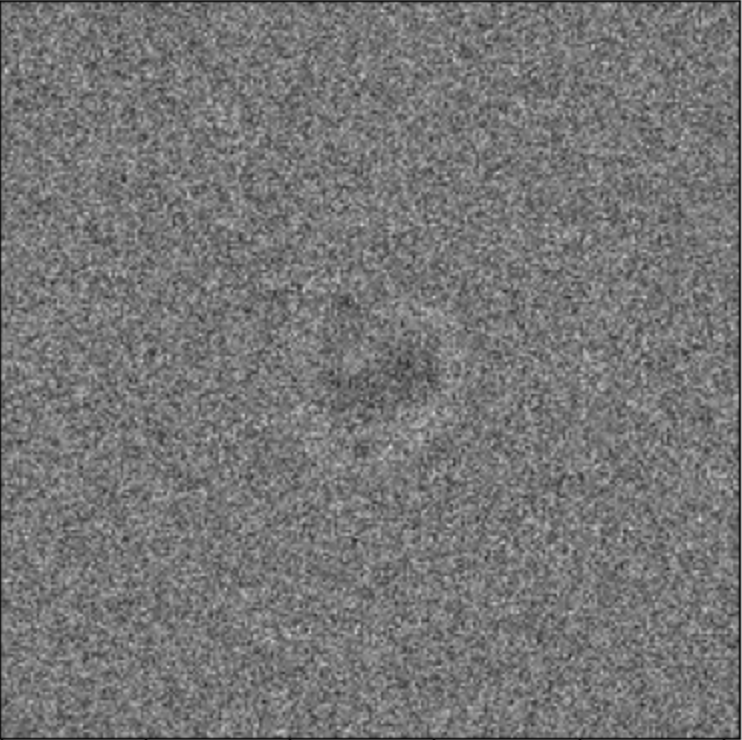}}
        \centerline{(b)}
    \end{minipage}
    \begin{minipage}[b]{0.24\linewidth}
        \centering
        \centerline{\includegraphics[trim=350 100 350 50, clip, width=0.64\linewidth]{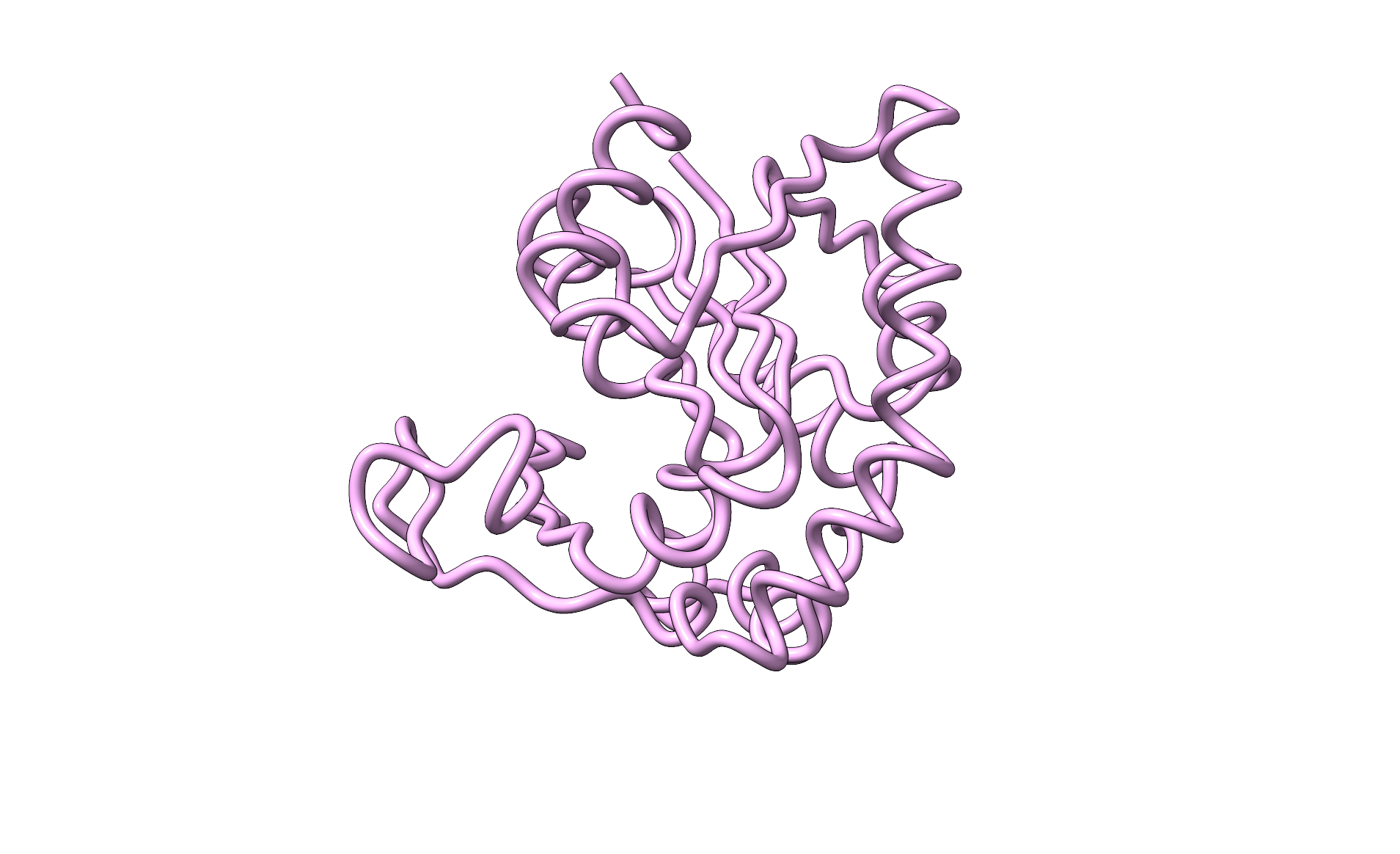}}
        \centerline{(c)}
    \end{minipage}
    \begin{minipage}[b]{0.24\linewidth}
        \centering
        \centerline{\includegraphics[trim=350 100 350 50, clip, width=0.64\linewidth]{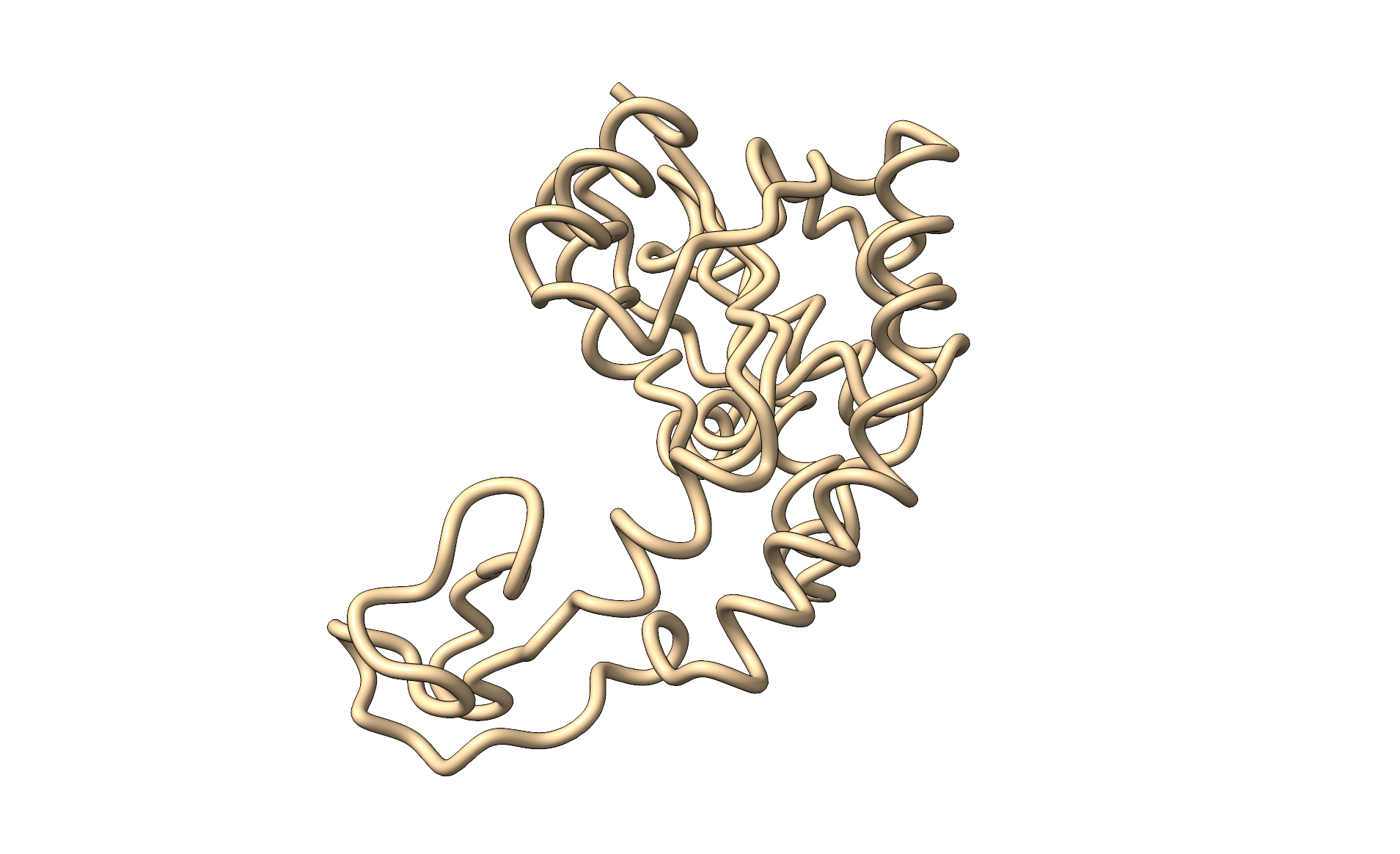}}
        \centerline{(d)}
    \end{minipage}
    \caption{Representative reconstruction. (a): Ground truth conformation. (b): Simulated cryo-EM image with randomized orientation. (c): Template conformation generated by AlphaFold 3, with a ground truth RMSD of 6.93 Å. (d): Final prediction by GNN autodecoder, with a ground truth RMSD of 1.85 Å.}
    \label{fig:visualized_results}
\end{figure*}

\subsection{Optimization}
The neural network weights $\theta$ and the latent variables $(z_i)_{i=1}^\ndata$ should minimize the objective $\loss \big(\theta,(z_i)_{i=1}^\ndata \big)$, given by
\begin{equation}
    \sum_{i=1}^\ndata \mathbb E_{\phi\sim\mu_i^\star}\|\datapoint_i-\fop(\phi.\nn_\theta(z_i))\|_2^2+\reg(\nn_\theta(z_i)).
\end{equation}
Here, $\mu_i^\star$ is the measure on $\operatorname{SO}(3)$ obtained from the ESL method applied to the conformation $\nn_\theta(z_i)$ and data $\datapoint_i$, and $\reg\colon\reconspace\to\Rpos$ is a regularization term acting on the predicted conformations.

Our regularization acts on each conformation separately, and is divided into three terms, $\reg=\regparam_0\reg_0+\lambda_1\reg_1+\lambda_2\reg_2$, where $\regparam_0,\regparam_1,\regparam_2\in\Rpos$ are regularization parameters.
The first term, $\reg_0$, is given by the squared norm of the mean of the 3D coordinates in $x$.
This term prevents reconstructed conformations from moving too much off center.
The second term enforces the preservation of interatomic distances along the backbone, and is given by
\begin{equation}
    \reg_1(\struct)=\frac{1}{\natoms}\sum_{i=1}^{\natoms-1} \left(\|x^{i+1}-x^i\|_2-\|\template^{i+1}-\template^i\|_2\right)^2.
\end{equation}
The third term is based on the work by Diepeveen et al., 2024~\cite{Diepeveen2024, krook2025}, and is given by
\begin{equation}
    \reg_2(\struct)=\sum_{i=1}^\natoms\sum_{j=i+1}^\natoms e^{-\omega|i-j|}\left(\log\left(\frac{\|\struct^i-\struct^j\|_2}{\|\template^i-\template^j\|_2}\right)\right)^2.
\end{equation}
This term punishes atoms that come too close to each other, and is more flexible than $\reg_1$ since taking logarithms makes the value increase of $\reg_2$ diminishes when distances grow.
The parameter $\omega\geq0$ controls how fast the rigidity of the protein backbone decreases along the chain.

The model weights and latent variables are initialized with random values and optimized using the Adam algorithm across mini-batches, covering all datapoints in every epoch~\cite{Kingma2014AdamAM}. The complete optimization procedure with ESL is described in Algorithm~\ref{alg:training} (with known poses, steps 7--10 are replaced with a single data discrepancy calculation). For additional details, see the supplemental materials.

\section{Experimental Validation}

In order to validate our approach, we generated synthetic cryo-EM datasets with ground truth conformations and poses, allowing us to compare different reconstruction methods. 

\subsection{Metrics}

As our validation metric we compute the root mean squared deviation~(RMSD) between each of the predicted conformations and the ground truth, measured in Ångström~(Å).
Since conformations are only determined up to pose, the RMSD is defined as the minimum over all relative poses.
As the aggregated metric we report the average RMSD across each dataset. For reference, the average RMSD of reconstructions is limited to $\sim\!\!1$~Å, partly due to the coarse forward model and the inherent noise in the data.

\subsection{Datasets}

To generate appropriate datasets we obtained simulated molecular dynamics trajectories for two proteins. The first trajectory features the 214-residue protein adenylate kinase (ADK) undergoing a closed-to-open transition~\cite{Beckstein2018}, and is discretized into 102 distinct conformations. The second trajectory features the 590-residue SARS-CoV-2 NSP-13 protein (NSP) at steady state~\cite{sarscov}, and is discretized into 200 distinct conformations. These trajectories are sufficiently heterogeneous to ensure that a consensus reconstruction cannot match all conformational states, as indicated by Fig.~\ref{fig:confs}. 

We then employed the transmission electron microscopy simulator Parakeet~\cite{parakeet2021} to generate 1~000 single-particle images for each of the conformations, applying uniformly random orientations and centered positions. In particular, the electron dose was set to 100 electrons/Å$^2$, comparable to realistic doses. For additional parameters, see the supplemental materials. Note that Parakeet does not use the same forward model as the one described in Section~\ref{sec:forward}, instead relying on a more accurate multi-slice method. The resulting datasets consist of 102~000 and 200~000 single-particle images for ADK and NSP, respectively. Access to the corresponding ground truth conformations allow us to evaluate the accuracy of reconstructed conformations in terms of per-image RMSD values. As template conformations we obtained independent predictions from AlphaFold~3~\cite{Abramson2024}. 

\subsection{Results}

With the datasets in place, the various conformations were reconstructed using different autodecoder architectures. We compared our proposed GNN architecture to a standard MLP architecture of comparable size, with $\sim\!\!30$k weights for ADK and $\sim\!\!150$k weights for NSP. In addition, we also compared outcomes with and without $\reg_2$-regularization. Both datasets were reconstructed with known poses, while ESL pose estimation was performed only for the ADK dataset. The dimension of the latent variables $z_i$ was set to 8 in all experiments. For additional details, see the supplemental materials.


To optimize the models we employed 4 Nvidia H200 GPUs, reaching convergence for ADK after 15 minutes with known poses and after 48 hours with ESL, while NSP required 24 hours with known poses. Each experiment was repeated using three different seeds. The RMSD averages are reported in Table~\ref{tab:rmsd_avgs}, a histogram of RMSD values for the ADK dataset before and after optimizing the GNN autodecoder with ESL and $\reg_2$-regularization is featured in Fig.~\ref{fig:histogram}, and a representative reconstruction is visualized in Fig.~\ref{fig:visualized_results}.

\begin{table}[t] 
    \caption{Final RMSD averages (Å).} 
    \label{tab:rmsd_avgs}
    \centering
    \vspace{0.2em}
    \begin{tabular}{c|ccc}
         & & With $\reg_2$ & Without $\reg_2$ \\
        \toprule
        \multirow{2}{*}{ADK (ESL)} & GNN & \textbf{1.92} & 2.21 \\
         & MLP & 1.95 & 2.94 \\
        \midrule
        \multirow{2}{*}{ADK (Known poses)} & GNN & \textbf{1.09} & 1.44 \\
         & MLP & 1.24 & 2.30 \\
        \midrule
        \multirow{2}{*}{NSP (Known poses)} & GNN & \textbf{2.08} & 2.38 \\
         & MLP & 2.37 & 3.32 \\
        \bottomrule
    \end{tabular}
\end{table}

Our experiments show promising results for the applicability of a geometry-aware GNN architecture, and that pairing it with the ESL method yields an effective solution to the heterogeneous atomistic 3D reconstruction problem in cryo-EM.
We note, however, that the $\reg_2$-regularization term does improve upon the MLP autodecoder results, suggesting that some of the regularizing effect of~$\reg_2$ is already provided by the inductive bias of the GNN architecture.
An interesting direction of future work is therefore to investigate whether more sophisticated topological neural networks can further improve reconstructions.
\section{Conclusion}
We introduced a GNN-based model for solving the cryo-EM 3D reconstruction problem under continuous heterogeneity, incorporating pose estimation with the ESL method and geometrically motivated regularization.
Results are promising when testing on synthetic data, showing improved reconstructions using GNNs over MLPs.
Further improvements may enable more accurate reconstructions of larger proteins, perhaps by using sophisticated geometric constructions than graphs.

\clearpage
\bibliographystyle{IEEEbib}
\bibliography{ref}
\end{document}


\maketitle

\section{Contrast transfer function}

The CTF function that we used for the forward model is a function of the Fourier-transformed coordinates in the plane, given by 
\begin{align*}
\label{eq:ctf}
    \ctf(\xi) =-A(\xi)\Bigg(&\!\sqrt{1-\alpha^2}\sin\left(\frac{\Delta z}{2k}|\xi|^2-\frac{C_s}{4k^3}|\xi|^4\right) \\
    &\,\,\,\,\quad+\alpha\cos\left(\frac{\Delta z}{2k}|\xi|^2-\frac{C_s}{4k^3}|\xi|^4\right)\!\Bigg).
\end{align*}
In line with the CTF parameter values used for Parakeet (see below), our experiments used the parameter values $C_s=2.7$mm and $\Delta z = -2.0 \mu$m, while~$\alpha=0.06$ to emulate the typical amplitude phase contrast for these values.

\section{Parakeet}

For both datasets the parameters for the simulated images were 100 electrons per Å$^2$, 300 keV electron energy, -2.0 $\mu$m defocus, 2.7 mm spherical aberration, 1.0 Å slice thickness, 1.0 Å pixel size, and an image size of 256 by 256 pixels. The particles were centered and their orientations were sampled uniformly from the Haar measure on SO(3). 

\section{Model hyperparameters}

The graph convolutional decoder architecture included a single fully connected linear embedding layer from latents to node embeddings, then $K$ layers of $C$ channels with node embeddings of dimension $N$ and skip connections, and finally a single linear decoding layer outputting residue coordinates, all with ReLU activations. The multilayer perceptron decoder architecture included $H$ fully connected hidden layers of dimension $D$, all with ReLU activations. See Table~\ref{tab:modelparams} for the hyperparameters used in experiments and the corresponding total weight counts. Results with increased layer counts or hidden dimension for MLP models were similar or worse.

\begin{table}[h] 
    \caption{Model hyperparameters for experiments} 
    \vspace{0.5em}
    \label{tab:modelparams}
    \centering
    \begin{tabular}{c|cccc|ccc}
        \multirow{2}{*}{Dataset} & \multicolumn{4}{c}{GNN} & \multicolumn{3}{c}{MLP} \\
         & $K$ & $C$ & $N$ & Weights & $H$ & $D$ & Weights \\
        \midrule
        ADK & 16 & 16 & 16 & 35k & 8 & 32 & 30k \\
        \midrule
        NSP & 16 & 16 & 32 & 172k & 8 & 64 & 148k \\
        \bottomrule
    \end{tabular}
\end{table}

\section{Regularization parameters}

For all experiments we set the regularization parameters to~$\lambda_0 = 0.0005$,~$\lambda_1 = 0.01$,~$\omega = 0.25$, and either~$\lambda_2 = 0.01$ or~$\lambda_2 = 0$, i.e., with or without~$\reg_2$-regularization. 

\section{Optimization details}
For the Adam optimizer we used a base learning rate of 0.001, and otherwise the default parameters given by PyTorch version 2.10. In addition, the learning rate was linearly scaled by $\min(epoch/20, 1)$ and otherwise kept constant. The batch size with known poses was 256 for the ADK dataset and 128 for the SCN dataset, while ESL constrained the batch size to 16 for the ADK dataset.

\section{ESL parameters}

For the ESL implementation, we took an approximately uniform grid of 14761 points on $SO(3)$ from the original paper. With reference to the notations in the original paper, the ESL parameters were~$J_0 = 15$ and $\eta = 2/3$. 

\section{Template conformations}

To obtain the template conformations, we accessed the freely available Alphafold 3 server (https://alphafoldserver.com/) and input the primary structure of the relevant protein, i.e., the amino residue chain. The C$_\alpha$-coordinates of the predicted conformation were employed as our template.